# Projection-free Online Learning

**Elad Hazan**                                                    EHAZAN@IE.TECHNION.AC.IL
Technion - Israel Inst. of Tech.

**Satyen Kale**                                                   SCKALE@US.IBM.COM
IBM T.J. Watson Research Center

## Abstract

The computational bottleneck in applying online learning to massive data sets is usually the projection step. We present efficient online learning algorithms that eschew projections in favor of much more efficient linear optimization steps using the Frank-Wolfe technique. We obtain a range of regret bounds for online convex optimization, with better bounds for specific cases such as stochastic online smooth convex optimization.

Besides the computational advantage, other desirable features of our algorithms are that they are parameter-free in the stochastic case and produce sparse decisions. We apply our algorithms to computationally intensive applications of collaborative filtering, and show the theoretical improvements to be clearly visible on standard datasets.

## 1. Introduction

In recent years the online convex optimization model has become a prominent paradigm for online learning. Within this paradigm, the Online Gradient Descent algorithm of Zinkevich (2003), and its close cousin Stochastic Gradient Descent, have been successfully applied to many problems in theory and practice.

While these algorithms are usually very efficient, a computational bottleneck that limits their applicability in several applications is the projection step in the algorithm. Specifically, whenever we take a step that causes the current iterate to leave the convex domain of interest, thus leading to an infeasible point, we project the point back into the domain in order to restore feasibility. This projection step implies finding the nearest point in the domain in $\ell_2$ distance, and in the general amounts to solving a convex quadratic program over the domain.

In many settings of practical interest, while solving convex quadratic programs is out of the question, *linear* optimization can be carried out efficiently. In this paper we give efficient online learning algorithms that replace the projection step with a linear optimization step for a variety of settings, as given in the following theorem:

**Theorem 1.1.** *There is an algorithm scheme for online convex optimization that performs one linear optimization over the convex domain per iteration, and with appropriate modifications for each setting, obtains the following regret bounds:*

|                 | Stochastic        | Adversarial   |
|-----------------|-------------------|---------------|
| *Smooth costs*  | $\tilde{O}(\sqrt{T})$ | $O(T^{3/4})$ |
| *Non-smooth costs* | $\tilde{O}(T^{2/3})$ | $O(T^{3/4})$ |

*Furthermore, in each iteration $t$, the algorithm maintains an explicit, efficiently sampleable distribution over at most $t$ boundary points with expectation equal to the current iterate.*

The above theorem entails several appealing advantages over the existing methods for online learning, as we detail below:

**Computational efficiency.** In several learning domains of interest projection steps are computationally very expensive. One prominent example is online collaborative filtering, where the domain of interest is the set of all positive semidefinite matrices of bounded trace. Projecting into this set amounts to computing the singular value decomposition (SVD) of the matrix to be projected, whereas linear optimization is finding the top singular vectors, a much more efficient operation.

**Parameter free.** Since the basic primitive of our algorithm is linear optimization rather than gradient steps and projections, our algorithms in the stochastic case are naturally parameter-free. In particular there is no parameter corresponding to the learning-rate. This makes it particularly easy to implement, since no parameter tuning is necessary.

**Efficient representation and Sparsity.** Another computational issue with standard projected gradient descent methods is more subtle: frequently the domain of interest is the convex hull of "integer" decision points in Euclidean space (for example, in the online shortest paths problem (Awerbuch & Kleinberg, 2008)). In such problems, points in the convex hull represent distributions over integer decision points, and to output a valid decision point we must be able to sample from such distributions. Typically, this requires being able to decompose points in the interior of the convex hull as explicit convex combinations of boundary points, and then sampling from the induced distribution. Such a decomposition may also require a lot of computational effort (for example, in the online shortest paths problem, this decomposition amounts to computing a flow decomposition in a network).

In contrast, our algorithm explicitly maintains a distribution over the vertices (or more generally boundary points) of the decision set, thereby eliminating the need for any further decomposition. In fact, in round $t$ the distribution is supported on at most $t$ boundary points, thus giving a form of sparsity.

## 1.1. Some Appropriate Convex Domains

In several interesting online learning scenarios the underlying decision sets do not admit "practically efficient" projections. We list several interesting examples of such decision sets below.

**Bounded Trace Norm Matrices.** The set of matrices with bounded trace norm is a common decision set for applications such as matrix completion and collaborative filtering. For example, consider the set $\mathcal{K}$ of $m \times n$ matrices of trace norm bounded by some parameter $\tau$. Computing the projection of a matrix $\mathbf{X}$ requires computing the SVD of $\mathbf{X}$, which requires $O(nm^2)$ time in general assuming $m \leq n$. Linear optimization over $\mathcal{K}$ amounts to computing the top singular vectors of the matrix defining the objective, which can be done much faster: typically, linear time in the number of non-zero entries in the matrix.

**Flow polytope.** Given a directed acyclic graph $G$ with $n$ nodes and $m$ edges with a specified source node $s$ and a sink node $t$, consider the set of all paths from $s$ to $t$. Any such path can be represented as a vector in $\mathbb{R}^m$ by its indicator vector over edges. Let $\mathcal{K}$ be the convex hull of all such path vectors. This can be equivalently described as the set $K$ of all unit flows from $s$ to $t$, which can be represented as a polytope in $\mathbb{R}^m$ with $O(m + n)$ linear inequalities. This set $\mathcal{K}$ arises with the online shortest paths problem (Awerbuch & Kleinberg, 2008). Computing the projection of a vector on $\mathcal{K}$ amounts to solving a quadratic program on this polytope, for which the most efficient algorithm known takes $O(n^{3.5})$ time. Linear optimization on $\mathcal{K}$ is much easier: it amounts to finding the shortest path from $s$ to $t$ given weights on the edges, and can be done in linear time using dynamic programming.

**Matroid polytope.** Given a matroid $M = (E, I)$, where $|E| = n$, any independent set $A \in I$ can be represented as a vector in $\mathbb{R}^n$ by its indicator vector. The matroid polytope $\mathcal{K}$ is defined to be the convex hull of all such indicator vectors of independent sets. This polytope can be defined using $O(2^n)$ linear inequalities. Computing a projection on $\mathcal{K}$ is therefore a difficult operation (although polynomial time, see (Nagano, 2007)). Linear optimization over $\mathcal{K}$ is very easy however: there is a simple greedy algorithm (see, e.g. Schrijver (2003)), amounting to sorting the coordinates of the objective vector, which solves it in $O(n \log(n))$ time.

**Rotations.** Consider the set of all $n \times n$ rotation matrices. These are orthogonal matrices of determinant 1. Let $\mathcal{K}$ be the convex hull of all rotation matrices. This set $\mathcal{K}$ arises in the online learning of rotations problem (Hazan et al., 2010). Computing the projection of a matrix on $\mathcal{K}$ is very difficult since there is no succinct description[1] of $\mathcal{K}$. However, linear optimization on $\mathcal{K}$ is a classic problem known as Wahba's problem (Wahba, 1965), and can be solved using one singular value decomposition. The only method of computing projections on $\mathcal{K}$ that we know of uses the ellipsoid method with linear optimization.

## 1.2. Discussion of Main Result

Our main result is that assuming it is possible to do *linear optimization* over the convex domain efficiently, we can obtain good algorithms (in the sense of obtaining sublinear regret) for online convex optimization over the domain using an online version of the classic Frank-Wolfe algorithm (Frank & Wolfe, 1956).

The above statement needs clarification: Zinkevich (2003) shows (via his Online Gradient Descent algorithm) that it is possible to do online convex optimization solving *one* quadratic program over the domain per step. Since quadratic optimization can be reduced to a polynomial number of linear optimizations via the ellipsoid algorithm, we can therefore do online convex optimization solving a *polynomial* number of linear programs over the domain per step. In contrast, we show (via our Online Frank-Wolfe algorithm) that it is possible to do online convex optimization solving *one linear* program over the domain per step. This yields immediate computational benefits.

---

[1] At least, none that we are aware of, other than (Sanyal et al., 2011) for the special case of 3 dimensions.

Another computational benefit comes from the fact that the algorithm automatically computes a distribution over boundary points for the iterates. In fact, if we simply want to sample from the distribution, then there is a very natural procedure for doing that: with a certain explicitly specified probability (viz. $t^{-a}$, see Algorithm 1), we replace the current boundary point with a new one that is computed in the current iteration in the linear optimization. This also automatically gives a *lazy* versions of the algorithm, in which the chosen decision point is updated very infrequently.

Our regret bounds are always sublinear, but not always optimal (with the exception of the stochastic, smooth case, where we obtain optimal regret bounds via our methods). Thus, theoretically we have slower convergence, in terms of number of iterations, to the optimal decision point, but the computational savings per iteration lead to a faster algorithm overall. This is validated by our experiments in Section 5.

### 1.3. Related Work

The closest work related to ours is that of Kalai & Vempala (2005). They give an algorithm (previously considered by Hannan (1957)) for online *linear* optimization which performs one linear optimization over the decision set in each iteration. The striking feature of their work is they are able to show optimal $O(\sqrt{T})$ regret bounds even for adversarial costs, although the limitation of their work is that the algorithm specifically works only for linear cost functions. They also give lazy versions of their algorithm via a careful correlation of randomness from one iteration to the next. In comparison, our algorithm has a very natural lazy implementation (simply replace the previous decision point with a new one with an explicitly specified probability) which, in our opinion, is significantly simpler.

Our results build upon the work of Clarkson (2010), Hazan (2008) and Jaggi (2011), who worked out the Frank-Wolfe technique for the problem of minimizing a single, static *smooth* convex cost function over a convex domain. We show how to extend their techniques to handle online, changing cost functions in stochastic and adversarial settings, and also show how to handle non-smooth functions.

## 2. Preliminaries

**Online convex optimization.** The problem of interest is online convex optimization (see the survey of Hazan (2011) for more details). Iteratively in each round $t = 1, 2, \ldots, T$ a learner is required to produce a point $\mathbf{x}_t$ from a convex, compact set $\mathcal{K} \subset \mathbb{R}^n$. In response, an adversary produces a convex cost function $f_t : \mathcal{K} \to \mathbb{R}$, and the learner suffers the cost $f_t(\mathbf{x}_t)$. The goal of the learner is to produce points $\mathbf{x}_t$ so that the regret,

$$\text{Regret} := \sum_{t=1}^{T} f_t(\mathbf{x}_t) - \min_{\mathbf{x} \in \mathcal{K}} \sum_t f_t(\mathbf{x}),$$

is sublinear in $T$. If the cost functions are stochastic, regret is measured using the expected cost function $f = \mathbf{E}[f_t]$ instead of the actual costs.

We assume that the set $\mathcal{K}$ diameter bounded by $D$ and it is possible to efficiently minimize a linear function, viz. computing $\arg\min_{\mathbf{x} \in \mathcal{K}} \mathbf{v} \cdot \mathbf{x}$ for some given vector $\mathbf{v} \in \mathbb{R}^n$ is easy. The cost functions $f_t$ are assumed to be $L$-Lipschitz, i.e. for any two points $\mathbf{x}, \mathbf{y} \in \mathcal{K}$, we have $|f_t(\mathbf{x}) - f_t(\mathbf{y})| \le L\|\mathbf{x} - \mathbf{y}\|$.

**Definition 2.1.** *Let $f : \mathbb{R}^n \mapsto \mathbb{R}$ be an arbitrary convex function which is also $L$-lipschitz.*
*$f$ is called $\beta$-smooth if for all $\mathbf{x}, \mathbf{y} \in \mathcal{K}$ we have*

$$f(\mathbf{x} + \mathbf{y}) \ \le \ f(\mathbf{x}) + \nabla f(\mathbf{x}) \cdot \mathbf{y} + \beta\|\mathbf{y}\|^2.$$

*$f$ is called $\sigma$-strongly convex if for all $\mathbf{x}, \mathbf{y} \in \mathcal{K}$ we have*

$$f(\mathbf{x} + \mathbf{y}) \ \ge \ f(\mathbf{x}) + \nabla f(\mathbf{x}) \cdot \mathbf{y} + \sigma\|\mathbf{y}\|^2.$$

Note that if $f$ is twice differentiable, then $\beta$ is upper bounded by the largest eigenvalue of the Hessian of $f$. The above definition together with first order optimality conditions imply that for a $\sigma$-strongly convex function $f$, if $\mathbf{x}^* = \arg\min_{\mathbf{x} \in \mathcal{K}} f(\mathbf{x})$, then

$$f(\mathbf{x}) - f(\mathbf{x}^*) \ \ge \ \sigma\|\mathbf{x} - \mathbf{x}^*\|^2. \qquad (1)$$

**Smoothed functions.** Let $\mathbb{B}$ and $\mathbb{S}$ denote the unit ball and unit sphere in $\mathbb{R}^n$ respectively. Given $\delta > 0$, let the $\delta$-smoothing of a function $f$ (c.f. (Flaxman et al., 2005)) be:

$$\hat{f}_\delta(\mathbf{x}) = \mathbf{E}_{\mathbf{u} \in \mathbb{B}}[f(\mathbf{x} + \delta\mathbf{u})],$$

where $\mathbf{u}$ is chosen uniformly at random from $\mathbb{B}$. We are implicitly assuming that $f$ is defined on all points within distance $\delta$ of $\mathcal{K}$. The following lemma (proof in the full version of this paper) shows that $\hat{f}_\delta$ is a good smooth approximation of $f$:

**Lemma 2.1.** *If $f$ is convex and $L$-Lipschitz, then the function $\hat{f}_\delta$ has the following properties:*

1. *$\hat{f}_\delta$ is convex and $L$-Lipschitz.*
2. *For any $\mathbf{x} \in \mathcal{K}$, $\nabla \hat{f}_\delta(\mathbf{x}) = \frac{d}{\delta} \mathbf{E}_{\mathbf{u} \in \mathbb{B}}[f(\mathbf{x} + \delta \mathbf{u})\mathbf{u}]$.*
3. *For any $\mathbf{x} \in \mathcal{K}$, $\|\nabla \hat{f}_\delta(\mathbf{x})\| \leq dL$.*
4. *$\hat{f}_\delta$ is $\frac{dL}{\delta}$-smooth.*
5. *For any $\mathbf{x} \in \mathcal{K}$, $|f(\mathbf{x}) - \hat{f}_\delta(\mathbf{x})| \leq \delta L$.*

**$\mathcal{K}$-Sparsity.** A feature of our algorithms is that they predict with sparse solutions, where sparsity is defined in the following manner.

**Definition 2.2.** *Let $\mathcal{K} \subseteq \mathbb{R}^n$ be a convex, compact set and let $\mathbf{x} \in \mathcal{K}$. We say that $\mathbf{x}$ is $t$-sparse w.r.t $\mathcal{K}$ if it can be written as a convex combination of $t$ boundary points of $\mathcal{K}$.*

All our algorithms produce $t$-sparse prediction at iteration $t$ w.r.t. the underlying decision set $\mathcal{K}$.

## 3. Algorithm and Analysis

### 3.1. Algorithm.

---
**Algorithm 1** Online Frank-Wolfe (OFW)
---
1: Input parameter: constant $a \geq 0$.
2: Initialize $\mathbf{x}_1$ arbitrarily.
3: **for** $t = 1, 2, \ldots, T$ **do**
4:    Play $\mathbf{x}_t$ and observe $f_t$.
5:    Compute $F_t = \frac{1}{t}\sum_{\tau=1}^{t} f_\tau$.
6:    Compute $\mathbf{v}_t \leftarrow \arg\min_{\mathbf{x} \in \mathcal{K}}\{\nabla F_t(\mathbf{x}_t) \cdot \mathbf{x}\}$.
7:    Set $\mathbf{x}_{t+1} = (1 - t^{-a})\mathbf{x}_t + t^{-a}\mathbf{v}_t$.
8: **end for**
---

### 3.2. Analysis.

Define $\Delta_t = F_t(\mathbf{x}_t) - F_t(\mathbf{x}_t^*)$, where $F_t = \frac{1}{t}\sum_{\tau=1}^{t} f_\tau$ as defined in step 1 of the algorithm, and $\mathbf{x}_t^* = \arg\min_{\mathbf{x} \in \mathcal{K}} F_t(\mathbf{x})$. The regret bounds all follow from the following general theorem:

**Theorem 3.1.** *Assume that for $t = 1, 2, \ldots, T$, the function $f_t$ is $L$-Lipschitz, $Bt^{-b}$-smooth for some constants $b \in [-1, 1/2]$ and $B \geq 0$, and $St^{-s}$-strongly convex for some constants $s \in [0, 1)$ and $S \geq 0$. Then in Algorithm 1, for all $t > 1$, we have*

$$\Delta_t \leq Ct^{-d},$$

*for both the following values of $C$ and $d$:*

$$(C, d) = \left(\max\{9D^2B,\ 3LD\}, \frac{1+b}{2}\right)$$

*and $(C, d) = \left(\max\{9D^2B,\ 36L^2/S,\ 3LD\}, \frac{2+2b-s}{3}\right)$.*

*In either case, this bound is obtained by setting $a = d - b$ in Algorithm 1.*

*Proof.* First, we note that $d \leq 1$ and $a \geq 0$ in either case since $b \in [-1, 1/2]$ and $s \in [0, 1)$, so that $t^{-a} \in [0, 1]$ and hence all iterates lie in $\mathcal{K}$. We prove the lemma by induction on $t$ for either values of $C$ and $d$. The statement is true for $t = 1$ since $f_1$ is $L$-Lipschitz, so $C \geq LD \geq L\|\mathbf{x}_1 - \mathbf{x}_1^*\| \geq f_1(\mathbf{x}_1) - f_1(\mathbf{x}_1^*) = \Delta_1$. So assume that for some $t \geq 1$, we have $\Delta_t \leq Ct^{-d}$. Now by convexity of $F_t$ we have

$$F_t(\mathbf{x}_t^*) \geq F_t(\mathbf{x}_t) + \nabla F_t(\mathbf{x}_t) \cdot (\mathbf{x}_t^* - \mathbf{x}_t)$$

Since $\mathbf{x}_t^* \in \mathcal{K}$, we have that $\nabla F_t(\mathbf{x}_t) \cdot \mathbf{x}_t^* \geq \nabla F_t(\mathbf{x}_t) \cdot \mathbf{v}_t$. From both observations:

$$\begin{aligned}
(\mathbf{v}_t - \mathbf{x}_t) \cdot \nabla F_t(\mathbf{x}_t) &\leq (\mathbf{x}_t^* - \mathbf{x}_t) \cdot \nabla F_t(\mathbf{x}_t) \\
&\leq F_t(\mathbf{x}_t^*) - F_t(\mathbf{x}_t) = -\Delta_t. \quad (2)
\end{aligned}$$

Since $f_t$ is $Bt^{-b}$-smooth, the smoothness of $F_t$ is bounded by $\frac{1}{t}\sum_{\tau=1}^{t} B\tau^{-b} \leq 3Bt^{-b}$ for $b \leq 1/2$. We now have

$$\begin{aligned}
F_t(\mathbf{x}_{t+1}) &= F_t(\mathbf{x}_t + t^{-a}(\mathbf{v}_t - \mathbf{x}_t)) \\
&\leq F_t(\mathbf{x}_t) + t^{-a}(\mathbf{v}_t - \mathbf{x}_t) \cdot \nabla F_t(\mathbf{x}_t) + 3D^2Bt^{-b-2a} \\
&\quad \text{(by } 3Bt^{-b}\text{-smoothness)} \\
&\leq F_t(\mathbf{x}_t) - t^{-a}\Delta_t + 3D^2Bt^{-b-2a},
\end{aligned}$$

using (2). Using the inequality $F_t(\mathbf{x}_t^*) \leq F_t(\mathbf{x}_{t+1}^*)$ in the bound above we get:

$$\begin{aligned}
F_t(\mathbf{x}_{t+1}) - F_t(\mathbf{x}_{t+1}^*) &\leq (1 - t^{-a})\Delta_t + 3D^2Bt^{-b-2a} \\
&\leq (1 - t^{-a})Ct^{-d} + 3D^2Bt^{-b-2a} \\
&\quad \text{(By induction hypothesis)} \\
&= Ct^{-d} - Ct^{b-2d} + 3D^2Bt^{b-2d} \\
&\quad \text{(Since } a = d - b) \\
&\leq Ct^{-d} - \frac{2}{3}Ct^{b-2d}, \quad (3)
\end{aligned}$$

since $C \geq 9D^2B$. In Lemma 3.1, we show the following bound using the strong convexity of the $f_t$ functions and the parameter choices:

$$f_{t+1}(\mathbf{x}_{t+1}) - f_{t+1}(\mathbf{x}_{t+1}^*) \leq \frac{2}{3}Ct^{1+b-2d}.$$

Multiplying (3) by $t$, adding the above bound, and dividing by $t + 1$, we get

$$F_{t+1}(\mathbf{x}_{t+1}) - F_{t+1}(\mathbf{x}_{t+1}^*) \leq \frac{t}{t+1}Ct^{-d} \leq C(t+1)^{-d},$$

since $d \leq 1$, thus completing the induction. $\square$

**Lemma 3.1.** *In the setup of Theorem 3.1, assuming $\Delta_t \leq Ct^{-d}$, we have*

$$f_{t+1}(\mathbf{x}_{t+1}) - f_{t+1}(\mathbf{x}_{t+1}^*) \leq \frac{2}{3}Ct^{1+b-2d}.$$

*Proof.* Since $f_t$ is $\sigma_t$-strongly convex, the strong convexity of $F_t$ is at least

$$\frac{1}{t}\sum_{\tau=1}^{t}\sigma_\tau \;=\; \frac{1}{t}\sum_{\tau=1}^{t}S\tau^{-s} \;\geq\; St^{-s},$$

since $\tau^{-s} \geq t^{-s}$ for all $\tau \leq t$. Thus, since $\mathbf{x}_t^* = \arg\min_{\mathbf{x}\in\mathcal{K}} F_t(\mathbf{x})$, we have by (1):

$$Ct^{-d} \;\geq\; \Delta_t \;=\; F_t(\mathbf{x}_t)-F_t(\mathbf{x}_t^*) \;\geq\; St^{-s}\|\mathbf{x}_t-\mathbf{x}_t^*\|^2,$$

which implies that $\|\mathbf{x}_t-\mathbf{x}_t^*\| \leq \sqrt{C/S}t^{s/2-d/2}$. By a similar argument, since by a simple calculation (details in the full version of this paper) we have $F_{t+1}(\mathbf{x}_t^*) - F_{t+1}(\mathbf{x}_{t+1}^*) \leq \frac{LD}{t+1}$, we conclude that

$$\|\mathbf{x}_t^*-\mathbf{x}_{t+1}^*\| \leq \sqrt{(LD/S)(t+1)^{s-1}} \leq \sqrt{C/S}t^{s/2-d/2},$$

since $C \geq 3LD$, $s < 1$ and $d \leq 1$. Thus, using the triangle inequality and the trivial bound $\|\mathbf{x}_t-\mathbf{x}_{t+1}^*\| \leq D$, we get

$$\|\mathbf{x}_t-\mathbf{x}_{t+1}^*\| \leq \min\{2\sqrt{(C/S)}t^{s/2-d/2}, D\} \leq \frac{C}{3L}t^{1+b-2d},$$

since $C \geq 36L^2/S$ if $d = \frac{2+2b-s}{3}$, and $C \geq 3LD$ if $d = \frac{1+b}{2}$. Furthermore, we have

$$\|\mathbf{x}_{t+1}-\mathbf{x}_t\| \;=\; t^{-a}\|\mathbf{v}_t-\mathbf{x}_t\|$$
$$\leq Dt^{-a} \;=\; Dt^{b-d} \;\leq\; \frac{C}{3L}t^{1+b-2d},$$

since $C \geq 3LD$ and $d \leq 1$. So by triangle inequality, we have

$$\|\mathbf{x}_{t+1}-\mathbf{x}_t^*\| \;\leq\; \frac{2C}{3L}t^{1+b-2d}.$$

Since $f_{t+1}$ is $L$-Lipschitz, we get the required bound. $\square$

# 4. Regret Bounds

## 4.1. Stochastic Costs

Assume now that the cost functions $f_t$ are sampled i.i.d. from some unknown distribution, and let $f^* = \mathbf{E}[f_t]$, and let $\mathbf{x}^* = \arg\min_{\mathbf{x}\in\mathcal{K}} f^*(\mathbf{x})$.

### 4.1.1. SMOOTH STOCHASTIC COSTS

**Theorem 4.1.** *For $\beta$-smooth stochastic convex loss functions $f_t$, there is an algorithm such that with probability at least $1-\delta$, its regret is bounded as follows :*

$$\sum_{t=1}^{T}f^*(\mathbf{x}_t)-f^*(\mathbf{x}^*) \;=$$
$$O((D^2\beta + LD)\sqrt{nT\log(nT/\delta)\log(T)}).$$

*Proof.* For $\beta$-smooth stochastic convex loss functions $f_t$, the algorithm is OFW applied to the functions $f_t$ with parameter settings that we specify now. First, $f_t$ is $\beta$-smooth, so we can set $B = \beta$ and $b = 0$. Since we make no assumptions about the strong convexity of $f_t$, so we can set $S = 0$, and $s = 0$. For these settings, the optimal values of the parameters are $d = \frac{1+b}{2} = 1/2$, $a = d - b = 1/2$, and $C = \max\{9D^2\beta, 3LD\}$. Thus, for all $t$, we have:

$$F_t(\mathbf{x}_t) - F_t(\mathbf{x}_t^*) \;\leq\; C/\sqrt{t}.$$

This implies that for the optimal point $\mathbf{x}^*$ we have

$$F_t(\mathbf{x}_t) - F_t(\mathbf{x}^*) \;\leq\; C/\sqrt{t}. \qquad (4)$$

In (Shalev-Shwartz et al., 2009), Theorem 5, the following is proved:

**Theorem 4.2.** *With probability at least $1-\delta$, for any $\mathbf{x}\in\mathcal{K}$ and for all $t = 1, 2, \ldots, T$ we have*

$$|F_t(\mathbf{x})-f^*(\mathbf{x})| \;\leq\; LD\sqrt{n\log(n/\delta)\log(t)/t}.$$

Using this theorem and (4), we conclude that with probability at least $1-\delta$, we have

$$f^*(\mathbf{x}_t)-f^*(\mathbf{x}^*)\leq C/\sqrt{t}+2LD\sqrt{n\log(nt/\delta)\log(t)/t}.$$

Summing up from $t=1$ to $T$, we get that the regret is $O(C\sqrt{nT\log(nT/\delta)\log(T)})$ with probability at least $1-\delta$. $\square$

### 4.1.2. NON-SMOOTH STOCHASTIC COSTS

**Theorem 4.3.** *For non-smooth stochastic convex loss functions $f_t$, there is an algorithm such with probability at least $1-\delta$, its regret is bounded as follows:*

$$\sum_{t=1}^{T}f^*(\mathbf{x}_t)-f^*(\mathbf{x}^*) \;=$$
$$O(\sqrt{n}LDT^{2/3} + LD\sqrt{nT\log(nT/\delta)\log(T)}).$$

*Proof.* For non-smooth stochastic convex loss functions $f_t$, the algorithm is OFW applied to the $\delta_t$-smoothing of $f_t$, i.e. the functions $\hat{f}_{t,\delta_t}$ for $\delta_t = \sqrt{n}Dt^{-1/3}$. The function $\hat{f}_{t,\delta_t}$ is $\frac{nL}{\delta_t} = (\sqrt{n}L/D)t^{1/3}$ smooth, so $B = \sqrt{n}L/D$ and $b = -1/3$. Since we make no assumptions about the strong convexity of $\hat{f}_{t,\delta_t}$, so we can set $S = 0$, and $s = 0$. For these settings, the optimal values of the parameters are $d = \frac{1+b}{2} = 1/3$, $a = d - b = 2/3$ and $C = \max\{9\sqrt{n}LD, 3LD\} = 9\sqrt{n}LD$. Thus for all $t$, we have:

$$\hat{F}_t(\mathbf{x}_t) - \hat{F}_t(\mathbf{x}_t^*) \;\leq\; 9\sqrt{n}LDt^{-1/3},$$

where $\hat{F}_t(\mathbf{x}) = \frac{1}{t}\sum_{\tau=1}^{t}\hat{f}_{\tau,\delta_\tau}(\mathbf{x})$. Thus, for the optimal point $\mathbf{x}^*$, we have

$$\hat{F}_t(\mathbf{x}_t) - \hat{F}_t(\mathbf{x}^*) \leq 9\sqrt{n}LDt^{-1/3}.$$

Since $|\hat{f}_t(\mathbf{x}) - f_t(\mathbf{x})| \leq \delta_t L = \sqrt{n}LDt^{-1/3}$, we have

$$|\hat{F}_t(\mathbf{x}) - F_t(\mathbf{x})| \leq \frac{1}{t}\sum_{\tau=1}^{t}\sqrt{n}LD\tau^{-1/3} \leq 3\sqrt{n}LDt^{-1/3}.$$

Using the above two bounds we get that

$$F_t(\mathbf{x}_t) - F_t(\mathbf{x}^*) \leq 15\sqrt{n}LDt^{-1/3}.$$

From this point, arguing as in the proof of Theorem 4.1, we conclude that the regret is $O(\sqrt{n}LDT^{2/3} + LD\sqrt{nT\log(nT/\delta)\log(T)})$ with probability at least $1 - \delta$. □

## 4.2. Adversarial Cost Functions

**Theorem 4.4.** *For adversarial cost functions $f_t$ (smooth or non-smooth), there is an algorithm that has the following regret bound. For any $\mathbf{x}^* \in \mathcal{K}$ we have:*

$$\sum_{t=1}^{T}f_t(\mathbf{x}_t) - f_t(\mathbf{x}^*) \leq 57LDT^{3/4}.$$

*Proof.* We apply the OFW algorithm to the functions $\hat{f}_t$ defined as follows. Suppose the algorithm plays point $\mathbf{x}_t$ in round $t$, and the adversary provides the cost function $f_t$. Define

$$\hat{f}_t(\mathbf{x}) = \nabla f_t(\mathbf{x}_t) \cdot \mathbf{x} + \sigma_t\|\mathbf{x} - \mathbf{x}_1\|^2,$$

where $\sigma_t = (L/D)t^{-1/4}$ and $\nabla f_t(\mathbf{x}_t)$ is a subgradient of $f_t$ at $\mathbf{x}_t$ such that $\|\nabla f_t(\mathbf{x}_t)\| \leq L$.

We now estimate the Lipschitz, smoothness, and strong convexity parameters for $\hat{f}_t$. First, $\nabla \hat{f}_t(\mathbf{x}) = \nabla f_t(\mathbf{x}_t) + (2L/D)t^{-1/4}(\mathbf{x} - \mathbf{x}_1)$. Since $f_t$ is $L$-Lipschitz, $\|\nabla f_t(\mathbf{x}_t)\| \leq L$, and $\|\mathbf{x} - \mathbf{x}_1\| \leq D$, impying that $\|\nabla \hat{f}_t(\mathbf{x})\| \leq 3L$, which implies that $\hat{f}_t$ is $3L$-Lipschitz. Next, note that

$$\begin{aligned}\hat{f}_t(\mathbf{x} + \mathbf{y}) - \hat{f}_t(\mathbf{x}) \\ = \nabla f_t(\mathbf{x}_t) \cdot \mathbf{y} + 2\sigma_t(\mathbf{x} - \mathbf{x}_1) \cdot \mathbf{y} + \sigma_t\|\mathbf{y}\|^2 \\ = \nabla \hat{f}_t(\mathbf{x}) \cdot \mathbf{y} + \sigma_t\|\mathbf{y}\|^2.\end{aligned}$$

Thus $\hat{f}_t$ is $(L/D)t^{-1/4}$-smooth, so we set $B = L/D$ and $b = 1/4$. Also it is $(L/D)t^{-1/4}$-strongly convex, so we set $S = L/D$ and $s = 1/4$. For these settings, the optimal values of the parameters are $d = \frac{2+2b-s}{3} = 3/4$, $a = d - b = 1/4$ and $C = \max\{9D^2B, 36L^2/S, 3LD\} = 36LD$. Thus by Theorem 3.1, for all $t$, we have:

$$\Delta_t = \hat{F}_t(\mathbf{x}_t) - \hat{F}_t(\mathbf{x}_t^*) \leq 36LDt^{-3/4}, \quad (5)$$

where $\hat{F}_t(\mathbf{x}) = \frac{1}{t}\sum_{\tau=1}^{t}\hat{f}_\tau(\mathbf{x})$, $\mathbf{x}_t^* = \arg\min_{\mathbf{x}\in\mathcal{K}}\hat{F}_t(\mathbf{x})$.

Kalai & Vempala (2005) prove that the "Be-The-Leader" algorithm has no regret. In particular, since the $\mathbf{x}_t^* = \arg\min_{\mathbf{x}\in\mathcal{K}}\frac{1}{t}\sum_{\tau=1}^{t}\hat{f}_\tau(\mathbf{x})$, we have, for any $\mathbf{x}^* \in \mathcal{K}$,

$$\sum_{t=1}^{T}\hat{f}_t(\mathbf{x}_t^*) \leq \sum_{t=1}^{T}\hat{f}_t(\mathbf{x}^*). \quad (6)$$

By strong convexity and since $\mathbf{x}_t^* = \arg\min_{\mathbf{x}\in\mathcal{K}}\hat{F}_t(\mathbf{x})$, we have

$$\hat{F}_t(\mathbf{x}_t) - \hat{F}_t(\mathbf{x}_t^*) \geq \sigma_t\|\mathbf{x}_t - \mathbf{x}_t^*\|^2.$$

This implies that

$$\|\mathbf{x}_t - \mathbf{x}_t^*\| \leq \sqrt{\Delta_t/\sigma_t} \leq 6Dt^{-1/4},$$

using (5). Next, since $\hat{f}_t$ is $3L$-Lipschitz we get

$$\forall t: \ \hat{f}_t(\mathbf{x}_t) \leq \hat{f}_t(\mathbf{x}_t^*) + 18LDt^{-1/4}.$$

Summing up from $t = 1$ to $T$, using the bound $\sum_{t=1}^{T}t^{-1/4} \leq 3T^{3/4}$ and (6), we get that

$$\sum_{t=1}^{T}\hat{f}_t(\mathbf{x}_t) - \hat{f}_t(\mathbf{x}^*) \leq 54LDT^{3/4},$$

which implies that

$$\begin{aligned}\sum_{t=1}^{T}\nabla f_t(\mathbf{x}_t) \cdot (\mathbf{x}_t - \mathbf{x}^*) \\ \leq 54LDT^{3/4} + \sum_{t=1}^{T}\sigma_t\|\mathbf{x}^* - \mathbf{x}_1\|^2 \leq 57LDT^{3/4},\end{aligned} \quad (7)$$

since $\sum_{t=1}^{T}\sigma_t = \sum_{t=1}^{T}t^{-1/4} \leq 3T^{3/4}$ and $\|\mathbf{x}^* - \mathbf{x}_1\|^2 \leq D^2$. By convexity of $f_t$, we have

$$f_t(\mathbf{x}_t) - f_t(\mathbf{x}^*) \leq \nabla f_t(\mathbf{x}_t) \cdot (\mathbf{x}_t - \mathbf{x}^*).$$

Plugging this into (7), we get the stated bound. □

## 5. Experiments

To evaluate the performance benefits of OFW over OGD, we experimented with a simple test application, viz. online collaborative filtering. This problem is the following. In each round, the learner is required to produce an $m \times n$ matrix $\mathbf{X}$ with trace norm (i.e. sum of singular values) bounded by $\tau$, a parameter. This matrix is to be interpreted as supplying by users $i \in [m]$ rating for each item $j \in [n]$. The adversary then chooses an entry $(i, j)$ and reveals the true rating for it, viz. $y \in \mathbb{R}$. The learner suffers the squared loss $(X(i, j) - y)^2$. The goal is to compete with the set of all $m \times n$ matrices of trace norm bounded by $\tau$. The offline version of this problem has been extensively studied in recent times, see e.g. Candès & Recht (2009), Jaggi & Sulovský (2010), Srebro et al. (2010),

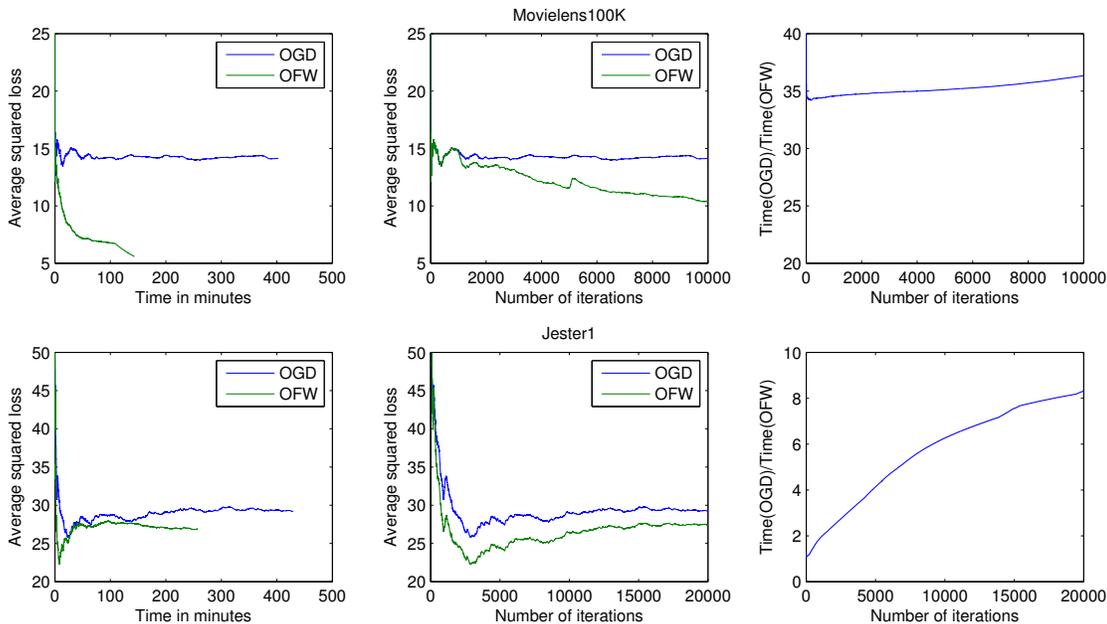

*Figure 1.* Comparison of OGD and OFW on the MovieLens100K (top row) and Jester1 (bottom row). The left plot shows average squared loss vs. running time for 100000 OFW iterations, and 10000 and 20000 OGD iterations for the two datasets respectively. The middle and right graphs show average squared loss and ratio of the running times of OGD and OFW respectively for as many iterations as OGD was run.

Lee et al. (2010), Salakhutdinov & Srebro (2010) and Shamir & Shalev-Shwartz (2011).

As mentioned previously, OGD requires computing the SVD of the matrix in each iteration, an $O(nm^2)$ time operation (assuming $m \leq n$), whereas OFW requires computing the top singular vector pair, an operation that in practice runs in near-linear time in the number of non-zero entries of the matrix.

**Datasets.** We used two publicly available datasets:

1. MovieLens100K (GroupLens): 100000 ratings in $\{1, 2, 3, 4, 5\}$ by 943 users on 1682 movies.
2. Jester1 (Goldberg et al., 2001): first 100000 ratings in $[-10, 10]$ by 24983 users on 100 jokes.

For simplicity, the sequence of entries $(i, j)$ chosen by the adversary is the same as the original sequence in these datasets. We also experimented with a $1000 \times 1000$ randomly generated matrix.

**Implementation.** We implemented the smooth, stochastic version of the OFW algorithm, even though the cost functions are *not* necessarily stochastic, mainly because of its faster convergence rate (in case of stochastic costs) and ease of implementation being parameter-free. We implemented the OGD and OFW algorithms in the most straightforward fashion in MATLAB, using the sparse matrices whenever possible, and the `svd` function for OGD and the `svds` function for OFW with a tolerance of $10^{-5}$. The running times were obtained on an 2.33GHz Intel® Xeon® CPU. All experiments were run in MATLAB in single-threaded mode using the `-singleCompThread` option. To ensure that the OFW and OGD runs completed in a reasonable amount of time (a few hours), we ran 100000 OFW iterations for both datasets, and only the first 10000 and 20000 OGD iterations for MovieLens100K and Jester1 respectively. The trace norm bounds used were 5000 and 200 respectively with no tuning.

**Results.** Figure 1 shows the results of our experiments. It can be clearly seen from the left plots that OFW is significantly faster than OGD, completing all its 100000 iterations much before the far fewer OGD iterations. Not only is it faster per iteration, surprisingly given that the costs are not stochastic, OFW also reduces the average squared loss faster than OGD, as can be seen from the middle plots. The right plots show that OFW is consistently around 35 times faster than OGD for the MovieLens100K dataset and around 6 times faster for the Jester1 dataset, and in fact this ratio keeps increasing as the number of iterations increases and the matrix parameter becomes more and more dense. This is reasonable since computing the SVD of a dense matrix requires much more effort than for a sparser matrix. The reason OGD is only about 6 times faster than OFW for Jester1 is because the matrix involved is tall-and-skinny, hav-ing only 100 columns, thus making the `svd` function almost as fast as the `svds` function. In our experiments with $1000 \times 1000$ randomly generated matrices we found that OFW was as much as 150 times faster than OGD.

## 6. Conclusions and Open Problems

In this paper, we gave an efficient algorithmic scheme for online convex optimization that performs one linear optimization per iteration rather than one quadratic optimization. The advantages over traditional gradient-descent techniques are speed of implementation, parameter-independence, explicit sampling scheme for iterates, sparsity, and natural lazy implementation. The disadvantage is that the provable regret bounds are not always optimal. The major open problem left is to improve the regret bounds, or show lower bounds on the number of linear optimizations necessary to obtain optimal regret with only one linear-optimization operation per iteration.